\documentclass[10pt,twocolumn,letterpaper]{article}

\usepackage{cvpr}
\usepackage{times}
\usepackage{epsfig}
\usepackage{graphicx}
\usepackage{amsmath}
\usepackage{amssymb}
\usepackage[hang,flushmargin]{footmisc} 

\usepackage[pagebackref=true,breaklinks=true,letterpaper=true,colorlinks,bookmarks=false]{hyperref}

\usepackage{setspace}
\cvprfinalcopy 

\ifcvprfinal\fi
\begin{document}

\title{$\mathbf{WRPN}$: Training and Inference using Wide Reduced-Precision Networks}

\author{
Asit Mishra \and Jeffrey J Cook \and Eriko Nurvitadhi \and Debbie Marr \\
   \\ 
   \hspace{-4.4in} Accelerator Architecture Lab, Intel Labs \vspace{-5mm} \\
}

\maketitle

\begin{abstract}
For computer vision applications, prior works have shown the efficacy of reducing numeric precision of model parameters (network weights) in deep neural networks but also that
reducing the precision of activations hurts model accuracy much more than reducing the precision of model parameters.
We study schemes to train networks from scratch using reduced-precision
activations without hurting the model accuracy. We reduce the precision
of activation maps (along with model parameters) using a novel 
quantization scheme and increase the number of filter maps in a layer, and find that this scheme compensates or surpasses the accuracy of the baseline full-precision network. As a result, one can significantly reduce the dynamic memory footprint, memory bandwidth, computational energy and speed up the training and inference process with appropriate hardware support. We call our scheme WRPN - wide reduced-precision networks. We report results using our proposed schemes and show that our results are better than previously reported accuracies on ILSVRC-12 dataset while being computationally less expensive compared to previously reported reduced-precision networks.
   
\end{abstract}

\section{Introduction}

Deep learning for robotics vision demands highly efficient real-time solutions. A promising approach to deliver one such extremely efficient solution is through the use of low numeric precision deep learning algorithms. Operating in lower precision mode reduces computation as well as data movement and storage requirements. 
Due to such efficiency benefits, there are many existing works that have proposed low-precision deep neural networks (DNNs), even down to 2-bit ternary mode~\cite{TTQ} and 1-bit mode~\cite{DoReFa, BNN}. However, the majority of existing works in low-precision DNNs sacrifice accuracy over the baseline full-precision networks. Further, most prior works target reducing the precision of the model parameters (network weights). This primarily benefits the inference step only when batch sizes are small.

We observe that activation maps (neuron outputs) occupy more memory compared to the model parameters for batch sizes typical during training. This observation holds even during inference when batch size is 8 or more for modern networks. Based on this observation, we study schemes for training and inference using low-precision DNNs where we reduce the precision of activation maps as well as the model parameters without sacrificing the network accuracy. We reduce the precision of activation maps (along with model parameters) and increase the number of filter maps in a layer. We call networks using this scheme wide reduced-precision networks (WRPN) and find that this scheme compensates or surpasses the accuracy of the baseline full-precision network. Although the number of raw compute operations increase as we increase the number of filter maps in a layer, the compute bits required per operation is now a fraction of what is required when using full-precision operations.

We present results using our scheme on AlexNet and ResNet on ILSVRC-12 dataset. Our results show that the proposed scheme offer better accuracies on ILSVRC-12 dataset, while being computationally less expensive compared to previously reported reduced-precision networks. Further, our reduced-precision quantization scheme is hardware friendly allowing for efficient hardware implementations of such networks for servers, deeply-embedded and real-time deployments.

\section{WRPN Scheme}

\begin{figure}[t]
\begin{center}
   \includegraphics[width=0.75\linewidth]{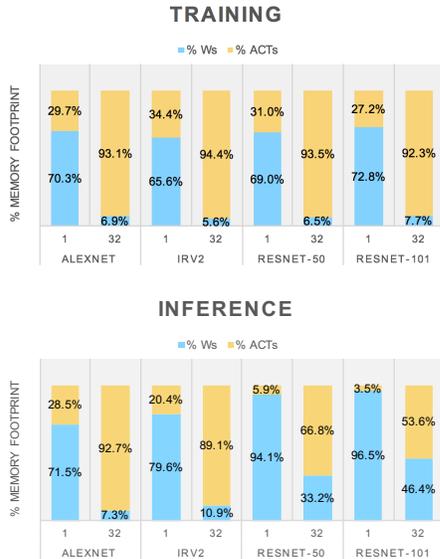}
\end{center}
   \caption{\small Memory footprint of activations and weights during training and inference for mini-batch sizes 1 and 32.}
\label{fig:MemoryFootprint}
\end{figure}

While most prior works proposing reduced-precision networks work with low precision weights (e.g.~\cite{BNN, TTQ, DoReFa, AALTWN}), we find that activation maps occupy a larger memory footprint when using mini-batches of inputs. 
Using mini-batches of inputs is typical in training of DNNs and for multi-modal inference~\cite{MULTIMODAL}. Figure~\ref{fig:MemoryFootprint} shows the memory footprint of activation maps and filter maps as batch size changes for 4 different networks during the training and inference steps. As batch-size increases, because of the filter reuse aspect across batches of inputs, the activation maps occupy significantly larger fraction of memory compared to the filter weights.

Based on this observation, we reduce the precision of activation maps for DNNs to speed up training and inference steps as well as cut down on memory requirements. However, a straightforward reduction in precision of activation maps leads to significant reduction in model accuracy. This has been reported in prior work as well~\cite{DoReFa}. We conduct a sensitivity study where we reduce the precision of activation maps and model weights for AlexNet network using ILSVRC-12 dataset. Table~\ref{table:AlexNet-accuracy} reports our findings. We find that reducing the precision of activation maps hurts model accuracy much more than reducing the precision of the filter parameters.

To re-gain the model accuracy while working with reduced-precision operands, we increase the \emph{number} of filter maps in a layer. Increasing the number of filter maps increases the layer compute complexity linearly. Another knob we tried was to increase the size of the filter maps. However increasing the filter map spatial dimension increases the compute complexity quadratically. Although the number of raw compute operations increase linearly as we increase the number of filter maps in a layer, the compute bits required per operation is now a fraction of what is required when using full-precision operations. As a result, with appropriate support, one can significantly reduce the dynamic memory requirements, memory bandwidth, computational energy and speed up the training and inference process. 

\begin{table}
\scriptsize	
\begin{center}
\begin{tabular}{|l||c|c|c|c|c|c|}
\hline	
Top-1 & 32b A & 8b A & 4b A & 2b A & 1b A \\
Acc.  & 1x/\textbf{2x} & 1x/\textbf{2x} & 1x/\textbf{2x} & 1x/\textbf{2x} & 1x/\textbf{2x} \\
\hline\hline
32b W	& 57.2/\textbf{60.5} & 54.3/\textbf{58.9} & 54.4/\textbf{58.6} & 52.7/\textbf{57.5} & -/\textbf{52.0} \\
8b W	& -/\textbf{-} & 54.5/\textbf{59.0} & 53.2/\textbf{58.8} & 51.5/\textbf{57.1} & -/\textbf{50.8}    \\
4b W	& -/\textbf{-}	& 54.2/\textbf{58.8} & 54.4/\textbf{58.6} & 52.4/\textbf{57.3} & -/\textbf{-}    \\
2b W	& 57.5/\textbf{-} & 50.2/\textbf{-} & 50.5/\textbf{57.2} & 51.3/\textbf{55.8} & -/\textbf{-} \\
1b W	& 56.8/\textbf{-} & -/\textbf{-} & -/\textbf{-}	& -	& 44.2/\textbf{-}      \\
\hline
\end{tabular}
\end{center}
\caption{\small AlexNet on ILSVRC-12 Top-1 accuracy \% as precision of
    activations (A) and weights (W) changes. - is a data-point we did
    not experiment for. Bold numbers are top-1 accuracy with 2x filters.
    All results are with end-to-end training of the network from
    scratch. 
    Noteworthy data-point is the comparison of 32b W and 32b A with 1x filter (baseline; 57.2\%) with 4b W and 4b A with 2x filter (58.6\%).}
\label{table:AlexNet-accuracy}
\end{table}

Table~\ref{table:AlexNet-accuracy} also reports the accuracy of AlexNet when we double the number of filter maps in a layer. We find that with doubling of filter maps, AlexNet with 4b weights and 2b activations exhibits accuracy at-par with full-precision networks. Operating with 4b weights and 4b activations, surpasses the baseline accuracy by 1.44\%. When doubling the number of filter maps, AlexNet's raw compute operations grow by 3.9x compared to the baseline full-precision network, however by using reduced-precision operands the overall compute complexity is a fraction of the baseline. For example, with 4b operands for weights and activations and 2x the number of filters, reduced-precision AlexNet is just 14\% of the
total compute cost\footnote{
\noindent Compute cost is the product of the number of FMA operations
and the width of the activation and weight operands.}  
of the full-precision baseline.

We also study how our scheme applies to bigger networks. For this we study widening the number of filters in ResNet-34. When doubling the 
number of filters in ResNet-34, we find that 4b weights and 8b
activations surpass the baseline accuracy (baseline full-precision
ResNet-34 is 73.2\% top-1 accuracy and our reduced-precision ResNet-34
is 73.8\% top-1 accuracy). The reduced-precision ResNet is 15.1\% the
compute cost of full-precision ResNet. 

\subsection{Hardware friendly quantization scheme}

For quantizing tensor values, we first constrain the weight values (W) to lie within the range \{-1,+1\} and activation values (A) within the range \{0,1\}. This is followed by a k-bit quantization of the values within the respective interval. Since a k-bit number can represent $2^{k}$ numbers, we represent the quantized weight values as $\frac{round((2^{k-1} - 1) * W)}{2^{k-1} - 1}$ and the quantized activation values as $\frac{round((2^{k} - 1) * A)}{2^{k-1}}$. One bit is reserved for sign-bit in case of weight values hence the use of $2^{k-1}$ for these quantized values. With appropriate affine transformations, the convolution operations (the bulk of the compute operations in the network) can be done using quantized values followed by scaling with floating-point constants. This makes our scheme hardware friendly and can be used in embedded systems. This is the quantization scheme for all the results reported above for AlexNet and ResNet.

\section{Conclusions}
Vision, speech and NLP based applications have seen tremendous success with DNNs. However, DNNs are compute intensive workloads. In this paper, we present WRPN scheme to reduce the compute requirements of DNNs. While most prior works look at reducing the precision of weights, we find that activations contribute significantly more to the memory footprint than weight parameters when using mini-batches and thus aggressively reduce the precision of activation values. Further, we increase the number of filter maps in a layer and show that widening of filters and reducing precision of network parameters does not sacrifice model accuracy. Our scheme is hardware friendly making it viable for deeply embedded system deployments which is useful for robotics applications.

{
\scriptsize
\bibliographystyle{ieee}
\bibliography{egbib}
}

\end{document}